# Local search for stable marriage problems[1]

M. Gelain and M. S. Pini and F. Rossi and K. B. Venable and T. Walsh

**Abstract**

The stable marriage (SM) problem has a wide variety of practical applications, ranging from matching resident doctors to hospitals, to matching students to schools, or more generally to any two-sided market. In the classical formulation, $n$ men and $n$ women express their preferences (via a strict total order) over the members of the other sex. Solving a SM problem means finding a stable marriage where stability is an envy-free notion: no man and woman who are not married to each other would both prefer each other to their partners or to being single. We consider both the classical stable marriage problem and one of its useful variations (denoted SMTI) where the men and women express their preferences in the form of an incomplete preference list with ties over a subset of the members of the other sex. Matchings are permitted only with people who appear in these preference lists, an we try to find a stable matching that marries as many people as possible. Whilst the SM problem is polynomial to solve, the SMTI problem is NP-hard. We propose to tackle both problems via a local search approach, which exploits properties of the problems to reduce the size of the neighborhood and to make local moves efficiently. We evaluate empirically our algorithm for SM problems by measuring its runtime behaviour and its ability to sample the lattice of all possible stable marriages. We evaluate our algorithm for SMTI problems in terms of both its runtime behaviour and its ability to find a maximum cardinality stable marriage. Experimental results suggest that for SM problems, the number of steps of our algorithm grows only as $O(n \log(n))$, and that it samples very well the set of all stable marriages. It is thus a fair and efficient approach to generate stable marriages. Furthermore, our approach for SMTI problems is able to solve large problems, quickly returning stable matchings of large and often optimal size despite the NP-hardness of this problem.

## 1 Introduction

The stable marriage problem (SM) [6] is a well-known problem of matching men to women to achieve a certain type of "stability". Each person expresses a strict preference ordering over the members of the opposite sex. The goal is to match men to women so that there are no two people of opposite sex who would both rather be matched with each other than with their current partners. The stable marriage problem has a wide variety of practical applications, ranging from matching resident doctors to hospitals, sailors to ships, primary school students to secondary schools, as well as in market trading. Surprisingly, such a stable marriage always exists and one can be found in polynomial time. Gale and Shapley give a quadratic time algorithm to solve this problem based on a series of proposals of the men to the women (or vice versa) [2].

There are many variants of the traditional formulation of the stable marriage problem. Some of the most useful in practice include incomplete preference lists (SMI), that allow one to model unacceptability for certain members of the other sex, and preference lists with ties (SMT), that model indifference in the preference ordering. With a SMI problem, the goal is to find a stable marriage in which the married people accept each other. It is known that all solutions of a SMI problem have the same size (that is, number of married people). In SMT problems, instead, solutions are stable marriages where everybody is married. Both of these variants are polynomial to solve. In real world situations, both ties and incomplete preference lists may be needed. Unfortunately, when we allow both, the problem becomes NP-hard [12]. In a SMTI (Stable Marriage with Ties and Incomplete lists) problem, there may be several stable marriages of different sizes, and solving the problem means finding a stable marriage of maximum size.

[1]Research partially supported by the Italian MIUR PRIN project 20089M932N: "Innovative and multi-disciplinary approaches for constraint and preference reasoning".

In this paper we investigate the use of a local search approach to tackle both the classical and the NP-hard variant of the problem. In particular, when we consider the classical problem, we investigate the fairness of stable marriage procedures based on local search. On the other hand, for SMTI problems, we focus on efficiency. Our algorithms are based on the same schema: they start from a randomly chosen marriage and, at each step, we move to a neighbor marriage by minimizing the distance to stability, which is measured by the number of unstable pairs. To avoid redundant computation due to the possibly large number of unstable pairs, we consider only those that are undominated, since their elimination maximises the distance to stability. Random moves are also used, to avoid stagnation in local minima. The algorithms stop when they find a solution or when a given limit on the number of steps is reached. A solution for an SMTI is a perfect matching (that is, a stable marriage with no singles), whereas, for an SM, a solution is just a stable marriage.

For the SM problem, we performed experiments on randomly generated problems with up to 500 men and women. It is interesting to notice that our algorithm always finds a stable marriage. Also, its runtime behaviour shows that the number of steps grows as little as $O(n \log(n))$. We also tested the fairness of our algorithm at generating stable marriages, measuring how well the algorithm samples the set of all stable marriages. As it is non-deterministic, it should ideally return any of the possible stable marriages with equal probability. We measure this capability in the form of an entropy that should be as close to that of an uniform sampe as possible. The computed entropy is about 70% of that of an uniform sample, and even higher on problems with small size.

For the SMTI problem, we performed experiments on randomly generated problems of size 100. We observe that our algorithm is able to find stable marriages with at most two singles on average in tens of seconds at worst. The SMTI problem has been tackled also in [4], where the problem is modeled in terms of a constraint optimization problem and solved employing a constraint solver. This systematic approach is guaranteed to find always an optimal solution. However, our experimental results show that our local search algorithm in practice always appears to find optimal solutions. Moreover, it scales well to sizes much larger than those considered in [4]. An alternative approach to local search is to use approximation methods. An overview of some results on SM problems is presented in [3].

# 2 Background

In this section we give some basic notions about the stable marriage problem. In addition, we present some basic notions about local search.

## 2.1 Stable marriage problem

A stable marriage (SM) problem [6] consists of matching members of two different sets, usually called men and women. When there are $n$ men and $n$ women, the SM problem is said to have size $n$. Each person strictly ranks all members of the opposite sex. The goal is to match the men with the women so that there are no two people of opposite sex who would both rather marry each other than their current partners. If there are no such pairs (called blocking pairs) the marriage is *"stable"*.

**Definition 1 (Marriage)** *Given an SM $P$ of size $n$, a marriage $M$ is a one-to-one matching of the men and the women. If a man $m$ and a woman $w$ are matched in $M$, we write $M(m) = w$ and $M(w) = m$.*

**Definition 2 (Blocking pair)** *Given a marriage $M$, a pair $(m, w)$, where $m$ is a man and $w$ is a woman, is a blocking pair iff $m$ and $w$ are not partners in $M$, but $m$ prefers $w$ to $M(m)$ and $w$ prefers $m$ to $M(w)$.*

**Definition 3 (Stable Marriage)** *A marriage $M$ is stable iff it has no blocking pairs.*

A convenient and widely used SM representation is showed in Table 1, where each person is followed by his/her preference list in decreasing order.

| men's preference lists | women's preference lists |
|---|---|
| 1: 5 7 1 2 6 8 4 3 | 1: 5 3 7 6 1 2 8 4 |
| 2: 2 3 7 5 4 1 8 6 | 2: 8 6 3 5 7 2 1 4 |
| 3: 8 5 1 4 6 2 3 7 | 3: 1 5 6 2 4 8 7 3 |
| 4: 3 2 7 4 1 6 8 5 | 4: 8 7 3 2 4 1 5 6 |
| 5: 7 2 5 1 3 6 8 4 | 5: 6 4 7 3 8 1 2 5 |
| 6: 1 6 7 5 8 4 2 3 | 6: 2 8 5 4 6 3 7 1 |
| 7: 2 5 7 6 3 4 8 1 | 7: 7 5 2 1 8 6 4 3 |
| 8: 3 8 4 5 7 2 6 1 | 8: 7 4 1 5 2 3 6 8 |

Table 1: An example of an SM of size 8.

For example, Table 1 shows that man 1 prefers woman 5 to woman 7 to woman 1 and so on. It is known that, at least one stable marriage exists for every SM problem. For a given SM instance, we can define a partial order relation on the set of stable marriages.

**Definition 4 (Dominance)** *Let $M$ and $M'$ be two stable marriages. $M$ dominates $M'$ iff every man has a partner in $M$ who is at least as good as the one he has in $M'$.*

Under the partial order given by the dominance relation, the set of stable marriages forms a distributive lattice [11]. Gale and Shapley give a polynomial time algorithm (GS) to find the stable marriage at the top (or bottom) of this lattice [2]. The top of such lattice is the male optimal stable marriage $M_m$, that is optimal from the men's point of view. This means that there are no other stable marriages in which each man is married with the same woman or with a woman he prefers to the one in $M_m$. The GS algorithm can also be used to find the female optimal stable marriage $M_w$ (that is the bottom of the stable marriage lattice), which is optimal from the women's perspective, by just replacing men with women (and vice versa) before applying the algorithm. A clear way to represent this lattice is a Hasse diagram representing the transitive reduction of the partial order relation. Figure 1 shows the Hasse diagram of the SM in Table 1.

A common concern with the standard Gale-Shapley algorithm is that it unfairly favors one sex at the expense of the other. This gives rise to the problem of finding "fairer" stable marriages. Previous work on finding fair marriages has focused on algorithms for optimizing an objective function that captures the happiness of both genders [7, 9]. A different approach is to investigate non-deterministic procedures that can generate a random stable marriage from the lattice with a distribution which is as uniform as possible.

In [1] the authors use a Markov chain approach to sample the stable marriage lattice. More precisely, the edges of the lattice dictate exactly how to formalize the moves to walk from one stable marriage to another one, so that there are at most a linear number of moves at each step, these are easily identifiable, and they form reversible moves that connect the state space and converge to the uniform distribution. Unfortunately, Bhatnagar et al. show that this random walk has an exponential convergence time, which would appear to suggest that the approach may not be feasible in practice.

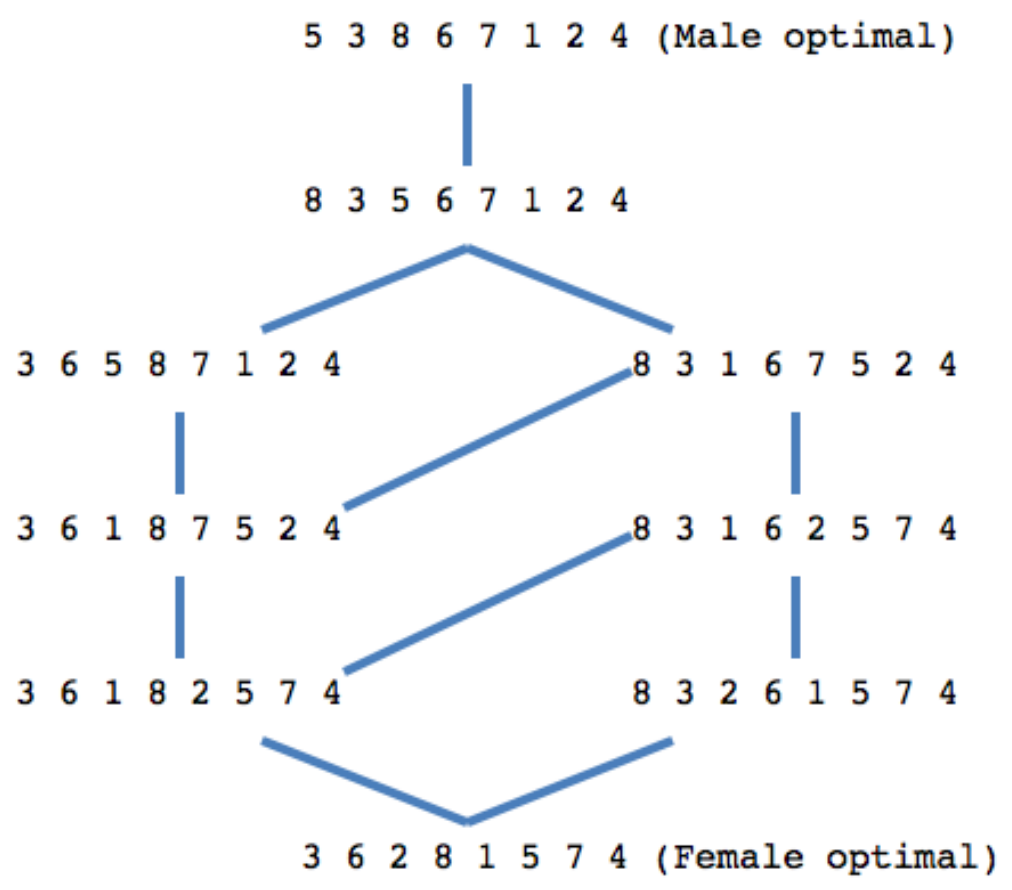

Figure 1: The Hasse diagram of the set of all stable marriages for the SM in Table 1.

In this paper we also consider a variant of the SM problem where preference lists may include ties and may be incomplete. This variant is denoted by SMTI [10]. Ties express indifference in the preference ordering, while incompleteness models unacceptability only for certain partners. Finally, our empirical data support the theoretical results in [14] about parameterized complexity of the stable marriage problem.

**Definition 5 (SMTI marriage)** *Given a SMTI problem with $n$ men and $n$ women, a marriage $M$ is a one-to-one matching between men and women such that partners accept each other. If a man $m$ and a woman $w$ are matched in $M$, we write $M(m) = w$ and $M(w) = m$. If a person $p$ is not matched in $M$ we say that he/she is single.*

**Definition 6 (Marriage size)** *Given a SMTI problem of size $n$ and a marriage $M$, its size is the number of men (or women) that are married.*

**Definition 7 (Blocking pairs in SMTI problems)** *Consider a SMTI problem $P$, a marriage $M$ for $P$, a man $m$ and a woman $w$. A pair $(m, w)$ is a blocking pair in $M$ iff $m$ and $w$ accept each other and $m$ is either single in $M$ or he strictly prefers $w$ to $M(m)$, and $w$ is either single in $M$ or she strictly prefers $m$ to $M(w)$.*

**Definition 8 (Weakly Stable Marriages)** *Given a SMTI problem $P$, a marriage $M$ for $P$ is weakly stable iff it has no blocking pairs.*

As we will consider only weakly stable marriages, we will simply call them stable marriages. Given a SMTI problem, there may be several stable marriages of different size. If the size of a marriage coincides with the size of the problem, it is said to be a perfect matching. Solving a SMTI problem means finding a stable marriage with maximal size. This problem is NP-hard [12].

## 2.2 Local search

Local search [8] is one of the fundamental paradigms for solving computationally hard combinatorial problems. Local search methods in many cases represent the only feasible way for solving large and complex instances. Moreover, they can naturally be used to solve optimization problems.

Given a problem instance, the basic idea underlying local search is to start from an initial search position in the space of all solutions (typically a randomly or heuristically generated candidate solution, which may be infeasible, sub-optimal or incomplete), and to improve iteratively this candidate solution by means of typically minor modifications. At each *search step* we move to a position selected from a *local neighborhood*, chosen via a heuristic evaluation function. The evaluation function typically maps the current candidate solution to a number such that the global minima correspond to solutions of the given problem instance. The algorithm moves to the neighbor with the smallest value of the evaluation function. This process is iterated until a *termination criterion* is satisfied. The termination criterion is usually the fact that a solution is found or that a predetermined number of steps is reached, although other variants may stop the search after a predefined amount of time.

Different local search methods vary in the definition of the neighborhood and of the evaluation function, as well as in the way in which situations are handled when no improvement is possible. To ensure that the search process does not stagnate in unsatisfactory candidate solutions, most local search methods use randomization: at every step, with a certain probability a random move is performed rather than the usual move to the best neighbor.

# 3 Local search on Stable Marriages

We now present an adaptation of the local search schema to deal with the classical stable marriage problem. Then, we will point out the aspects that have to be changed to deal with SMTI problems.

Given an SM problem $P$, we start from a randomly generated marriage $M$. Then, at each search step, we compute the set $BP$ of blocking pairs in $M$ and compute the neighborhood, which is the set of all marriages obtained obtained by removing one of the blocking pairs in $BP$ from $M$. Consider a blocking pair $bp = (m, w)$ in $M$, $m' = M(w)$, and $w' = M(m)$. Then, removing $bp$ from $M$ means obtaining a marriage $M'$ in which $m$ is married with $w$ and $m'$ is married with $w'$, leaving the other pairs unchanged. To select the neighbor $M'$ of $M$ to move to, we use an evaluation function $f : \mathcal{M}_n \rightarrow Z$, where $\mathcal{M}_n$ is the set of all possible marriages of size $n$, and $f(M) = nbp(M)$. For each marriage $M$, $nbp(M)$ is the number of blocking pairs in $M$, and we move to one with the smallest value of $f$.

To avoid stagnation in a local minimum of the evaluation function, at each search step we perform a random walk with probability $p$ (where $p$ is a parameter of the algorithm), which removes a randomly chosen blocking pair in $BP$ from the current marriage $M$. In this way we move to a randomly selected marriage in the neighborhood. The algorithm terminates if a stable marriage is found or when a maximal number of search steps or a timeout is reached.

This basic algorithm, called SML, has been improved in the computation of the neighborhood, obtaining SML1. When SML moves from one marriage to another one, it takes as input the current marriage $M$ and the list $PAIRS$ of its blocking pairs and returns the marriage in the neighborhood of $M$ with the best value of the evaluation function, i.e. the one with fewer blocking pairs. However, the number of such blocking pairs may be very large. Also, some of them may be useless, since their removal would surely lead to new marriages that will not be chosen by the evaluation function. This is the case for the so-called *dominated* blocking pairs. Algorithm SML1 considers only undominated blocking pairs.

**Definition 9 (Dominance in blocking pairs)** *Let* $(m, w)$ *and* $(m, w')$ *be two blocking pairs. Then* $(m, w)$ *dominates (from the men's point of view)* $(m, w')$ *iff* $m$ *prefers* $w$ *to* $w'$*.*

*There is an equivalent concept from the women's point of view.*

**Definition 10 (Undominated blocking pair)** *A men- (resp., women-) undominated blocking pair is a blocking pair such that there is no other blocking pair that dominates it from the men's (resp., women's) point of view.*

It is easy to see that, if $M$ is an unstable marriage, $(m, w)$ an men- (resp., women-) undominated blocking pair in $M$, $m' = M(w)$, $w' = M(m)$, and $M'$ is obtained from $M$ by removing $(m, w)$, there are no blocking pairs in $M'$ in which $m$ (resp., $w$) is involved. This property would not be true if we removed a dominated blocking pair. This is why we focus on the removal of undominated blocking pairs when we pass from one marriage to another in our local search algorithm.

Considering again the SM in Table 1 and the marriage 2 7 4 8 6 3 5 1. The blocking pair $(m_8, w_4)$ dominates (from the men's point of view) $(m_8, w_2)$. If we remove $(m_8, w_2)$ from the marriage, $(m_8, w_4)$ will remain. On the other hand, removing $(m_8, w_4)$ also eliminates $(m_8, w_2)$. Thus, removing $(m_8, w_4)$ is more useful than removing $(m_8, w_2)$.

By using the undominated blocking pairs instead of all the blocking pairs, we also limit the size of the neighborhood, since each man or woman is involved in at most one undominated blocking pair. Hence we have at most $2n$ neighbor marriages to evaluate.

Let us now analyse more carefully the set of blocking pairs considered by SML1. Consider the case in which a man $m_i$ is in two blocking pairs, say $(m_i, w_j)$ and $(m_i, w_k)$, and assume that $(m_i, w_j)$ dominates $(m_i, w_k)$ from the men's point of view. Then, let $w_j$ be in another blocking pair, say $(m_z, w_j)$, that dominates $(m_i, w_j)$ from the women's point of view. In this situation, SML1 returns $(m_z, w_j)$ because it computes the undominated blocking pairs from men's point of view (which are $(m_i, w_j)$ and $(m_z, w_j)$) and, among those, maintain the undominated ones from the women's point of view ($(m_z, w_j)$ in this case). The removal of $(m_z, w_j)$ automatically eliminates $(m_i, w_j)$ from the set of blocking pairs of the marriage, since it is dominated by $(m_z, w_j)$. However, the blocking pair $(m_i, w_k)$ is still present because the blocking pair that dominated it (i.e. $(m_i, w_j)$) is not a blocking pair any longer. We also consider a procedure that will return in addition the blocking pair $(m_i, w_k)$, so to avoid having to consider it again in the subsequent step of the local search algorithm. We call SML2 the algorithm obtained from SML1 by using this new way to compute the blocking pairs.

Since dominance between blocking pairs is defined from one gender's point of view, at the beginning of our algorithms we randomly choose a gender and, at each search step we change the role of the two genders. For example, in SML1, if we start by finding the undominated blocking pairs from the men's point of view and, among those, we keep only the undominated blocking pairs from the women's point of view, in the following second step we do the opposite, and so on. In this way we ensure that SML1 and SML2 are gender neutral.

Summarizing, we have defined three algorithms, called SML, SML1, and SML2, to find a stable marriage for a given SM instance. Such algorithms differ only for the set of blocking pairs considered to define the neighborhood.

# 4 Local search for SMTI problems

To adapt the SML algorithm to solve problems with ties and incomplete lists it is important to recall that an SMTI may have several stable marriages of different size. Thus, solving an SMTI problem means finding a stable marriage with maximal size. If the size of the marriage coincide with the size of the problem, it is said to be perfect and the algorithm can stop before the step limit. Otherwise the algorithm returns the best marriage found during search, defined as follows: if no stable marriage has been found, then the best marriage is the one with the smallest value of the evaluation function; otherwise, it is the stable marriage with fewest singles.

The SML algorithm is therefore modified in the following ways:

- the evaluation function has to take into account that some person may be not married, so we use: $f(M) = nbp(M) + ns(M)$, where, for each marriage $M$, $ns(M)$ is the number of singles in $M$ which are not in any blocking pair.
- When we remove a blocking pair $(m, w)$ from a marriage $M$, their partners $M(m)$ and $M(w)$ become single.
- The algorithm performs a random restart when a stable marriage is reached, since its neighborhood is empty (because it has no blocking pairs).

We call LTIU the modified algorithm for SMTI problems, obtained from SML by the above modifications and by using undominated blocking pairs.

# 5 Experiments

We tested our algorithms on randomly generated sets of SM and SMTI instances. For SM problems, we generated stable marriage problems of size $n$ using the impartial culture model (IC) [5] which assigns to each man and to each woman a preference list uniformly chosen from the $n!$ possible total orders of $n$ persons. This means that the probability of any particular ordering is $1/n!$.

For SMTI problems, we generated problems using the same method as in [4]. More precisely, the generator takes three parameters: the problem's size $n$, the probability of incompleteness $p_1$, and the probability of ties $p_2$. Given a triple $(n, p_1, p_2)$, a SMTI problem with $n$ men and $n$ women is generated, as follows:

1. For each man and woman, we generate a random preference list of size $n$, i.e., a permutation of $n$ persons;
2. We iterate over each man's preference list: for a man $m_i$ and for each women $w_j$ in his preference list, with probability $p_1$ we delete $w_j$ from $m_i$'s preference list and $m_i$ from $w_j$'s preference list. In this way we get a possibly incomplete preference list.
3. If any man or woman has an empty preference list, we discard the problem and go to step 1.
4. We iterate over each person's (men and women's) preference list as follows: for a man $m_i$ and for each woman in his preference list, in position $j \geq 2$, with probability $p_2$ we set the preference for that woman as the preference for the woman in position $j - 1$ (thus putting the two women in a tie).

Note that this method generates SMTI problems in which the acceptance is symmetric. If a man $m$ does not accept a woman $w$, $m$ is removed from $w$'s preference list as well. This does not introduce any loss of generality because $m$ and $w$ cannot be matched together in any stable marriage.

Notice also that this generator will not construct a SMTI problem in which a man (resp., woman) accepts only women (resp., men) who do not find him (resp, her) acceptable. Such a man (resp., woman) will remain single in every stable matching. A simple preprocessing step can remove such men and women from any problem, giving a smaller instance of the form constructed by our generator.

# 6 Results on SM problems

We measured the performance of our algorithms in terms of number of search steps. For these tests, we generated 100 SM problems for each of the following sizes: 100, 200, 300, 400 and 500. In the following we show only the results of our best algorithm, which is SML2. We studied how fast SML2 converges to a stable marriage, by measuring the ratio between the number of blocking pairs and the size of the problem during the execution. Figure 2(a) shows that SML2 has a very simple scaling behavior. Let us denote by $\langle b \rangle$ the average number of blocking pairs of the marriage found by SML2 for SM problems of size $n$ after $t$ steps. Then the experimental results shown in Figure 2(a) have a very good fit with the function $\langle b \rangle = an^2 2^{-bt/n}$, where $a$ and $b$ are constants computed empirically ($a \approx 0.25$ and $b \approx 5.7$). Figure 2(a) shows that the analytical function $\langle b \rangle$ has practically the same curve as the experimental data. The figure shows also that the average number of blocking pairs, normalized by dividing it by $n$, decreases during the search process in a way that is independent from the size of the problem.

We can use function $\langle b \rangle$ to conjecture the runtime behavior of our local search method. Consider the median number of steps, $t_{med}$, taken by SML2. Assume this occurs when half the problems have one blocking pair left and the other half have zero blocking pairs. Thus, $\langle b \rangle = \frac{1}{2}$. Substituting this value in the equation for $\langle b \rangle$, taking logs, solving for $t_{med}$, and grouping constant terms, we get $t_{med} = cn(d + 2\log_2(n))$ where $c$ and $d$ are constants. Hence, we can conclude that $t_{med}$ grows as $O(n \log(n))$.

We then fitted this equation for $t_{med}$ to the experimental data (using $c \approx 0.26$ and $d \approx -5.7$). The result is shown in Figure 2(b), where we see that the experimental data have the same curve as function $t_{med}$. This means that we can use such an equation to predict the number of steps our algorithms needs to solve a given SM instance.

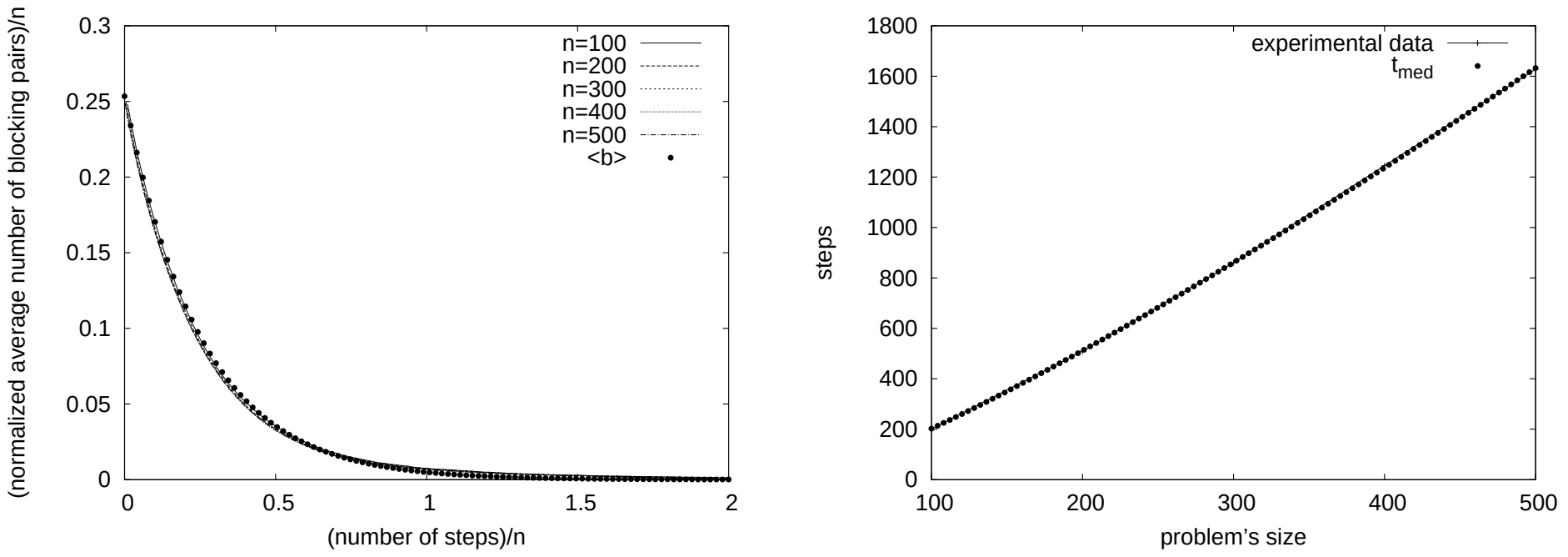

(a) Blocking pair ratio during the execution.

(b) Number of steps necessary to find a stable marriage.

Figure 2: Results using SML2.

## 6.1 Sampling the stable marriage lattice

We also evaluated the ability of SML2 to sample the lattice of stable marriages of a given SM problem. To do this, we randomly generated 100 SM problems for each size between 10 and 100, with step 10. Then, we run the SML2 algorithm 500 times on each instance. To evaluate the sampling capabilities of SML2, we first measured the distance of the found stable marriages (on average) from the male-optimal marriage (the one that would be returned by the GS algorithm).

Given a SM problem $P$, consider a stable marriage $M$ for $P$. The distance of $M$ from $M_m$ is the number of arcs from $M$ to $M_m$ in the Hasse diagram of the stable marriage lattice for $P$. This diagram can be computed in $O(n^2 + n|S|)$ time [7], where $S$ is the set of all possible stable marriages of a given SM instance. For each SM problem, we compute the average normalized distance from the male-optimal marriage considering 500 runs. Notice that normalizations is needed since different SM instances with the same size may have a different number of stable lattices. Then, we compute the average $D_m$ of these distances over all the 100 problems with the same size, which is therefore formally defined as $D_m = \frac{1}{100}\sum_{j=1}^{100}\frac{1}{500}\sum_{i=1}^{500}\frac{d_m(M_i,P_j)}{d_m(M_i,P_j)+d_w(M_i,P_j)}$, where $d_m(M_i, P_j)$ (resp., $d_w(M_i, P_j)$) is the distance of $M_i$ from the male (resp., female)-optimal marriage in the lattice of an SM $P_j$. If $D_m = 0$, it means that all the stable marriages returned coincides with the male-optimal marriage. On the other extreme, if $D_m = 1$, it means that all stable marriages returned coincide with the female-optimal one. Figure 3(a) shows that, for the stable marriages returned by algorithm SML2, the average distance from the male-optimal is around 0.5.

This is encouraging but not completely informative, since an algorithm which returns the same stable marriage all the times, with distance 0.5 from the male-optimal would also have $D_m = 0.5$. To have more informative results, we consider the entropy of the stable marriages returned by SML2. This measures the randomness in the solutions. Let $f(M_i)$ be the frequency that SML2 finds a marriage $M_i$ (for $i$ in $[1, |S|]$) that is: $f(M_i) = \frac{1}{500}\sum_{j=1}^{500}\mathbb{1}_{M_i}(j)$, where $\mathbb{1}_{M_i}(j)$ is the indicator function that returns 1 if in the $j$-th execution the algorithm finds $M_i$, and 0 otherwise. The entropy $E(P)$ for each SM instance $P$ (i.e., for each lattice) of size $m$ is then: $E(P) = -\sum_{i=1\in\{1..|S|\}} f(M_i)\log_2(f(M_i))$. In an ideal case, when each stable marriage in the lattice has a uniform probability of $1/m!$ to be reached, the entropy is $\log_2(|S|)$ bits. On the other hand, the worst case is when the same stable marriage is always returned, and the entropy is thus 0 bits. As we want a measure that is independent from the problem's size, we consider a normalized entropy, that is $E(P)/\log_2(|S|)$, which is in [0,1].

As we have 100 different problems for each size, we compute the average of the normalized entropies for each class of problems with the same size: $E_n = \frac{1}{100}\sum_{i=1}^{100} E(P_i)/\log_2(|S_i|)$, where $S_i$ is the set of stable marriages of $P_i$.

Figure 3(b) shows that SML2 is not far from the ideal behavior. The normalized entropy starts from a value of 0.85 per bit at size 10, decreasing to just above 0.6 per bit as the problem's size grows.

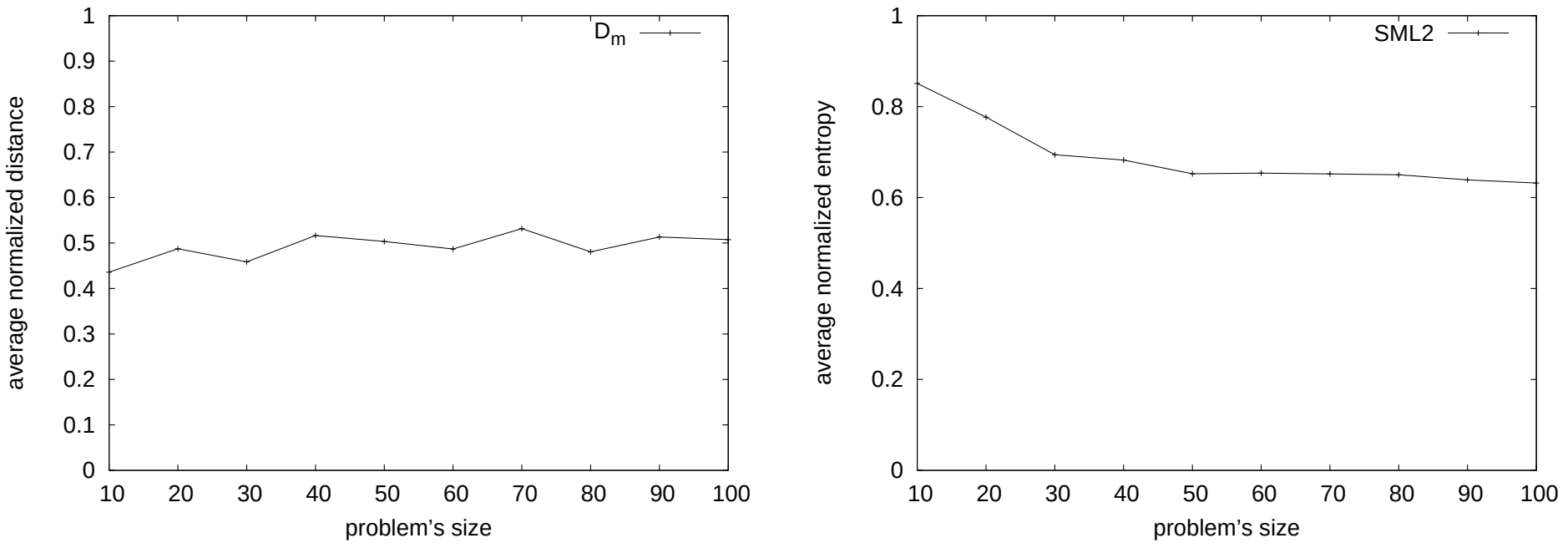

(a) Average normalized distance $D_m$ varying $n$.  (b) Entropy.

Figure 3: Sampling with SML2.

Considering both Figures 3(b) and 3(a), it appears that SML2 samples the stable marriage lattice very well. Considering also the distance $D_m$ (Figure 3(a)), the possible outcomes appear to be equally distributed along the paths from the top to the bottom of the lattice.

# 7 Results on SMTI problems

We generated random SMTI problems of size 100, by letting $p_2$ vary in [0, 1.0] with step 0.1, and $p_1$ vary in [0.1, 0.8] with step 0.1 (above 0.8 the preference lists start to be empty). For each parameter combination, we generated 100 problem instances. Moreover, the probability of the random walk is set to $p$=20% and the search step limit is $s$=50000.

We start by showing the average size of the marriages returned by LTIU. In Figure 4(a) we see that LTIU almost always finds a perfect marriage (that is, a stable marriage with no singles). Even in settings with a large amount of incompleteness (that is, $p1$ = 0.7 - 0.8) the algorithm finds very large marriages, with only 2 singles on average.

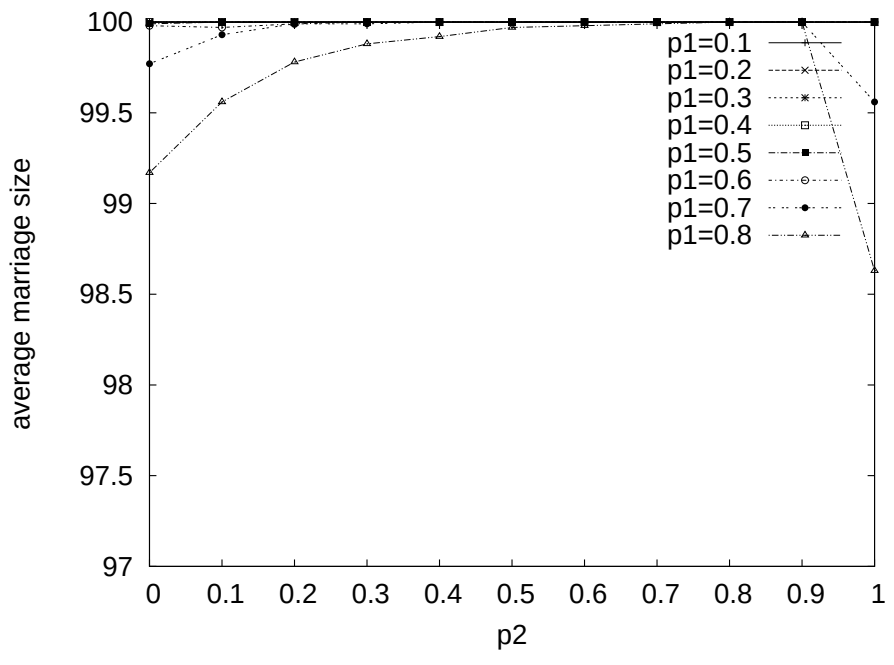

(a) Average size of marriages.

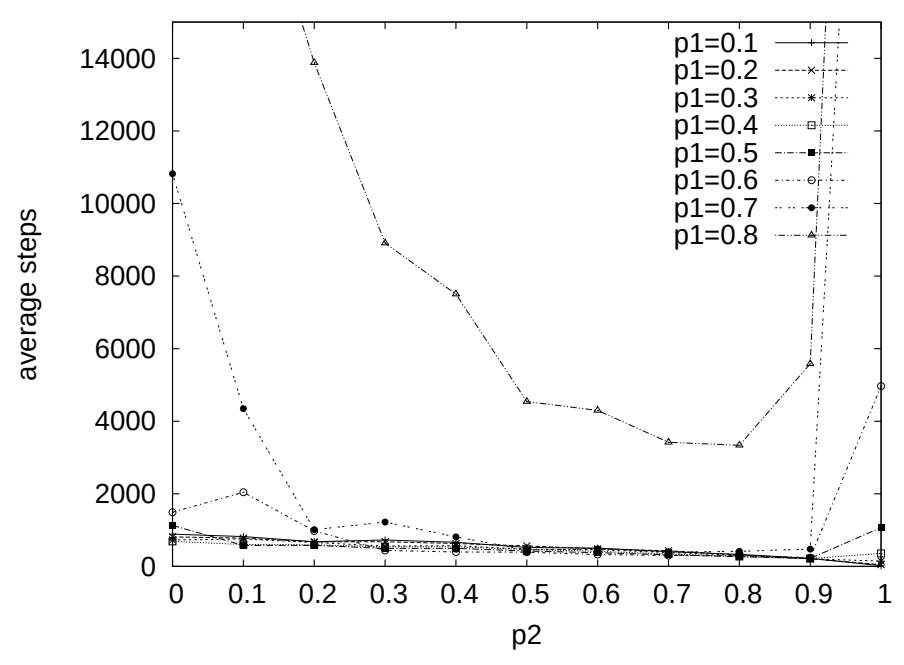

(b) Average number of steps.

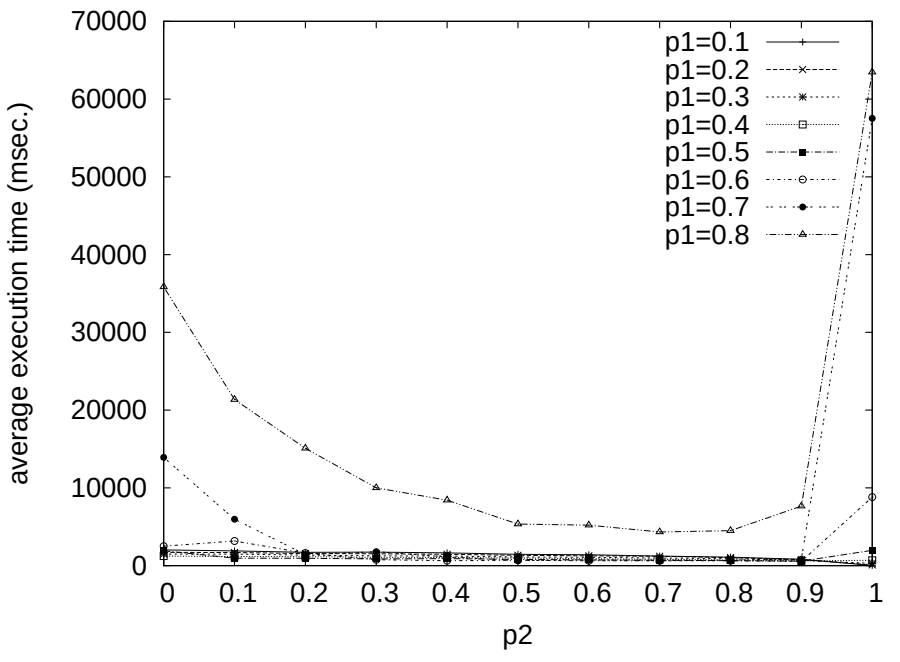

(c) Average execution time.

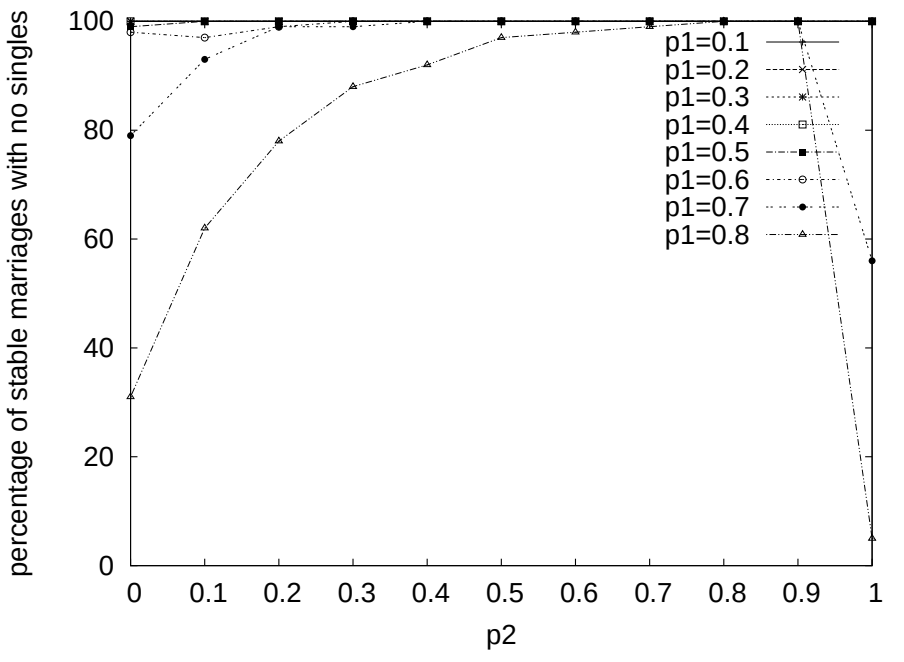

(d) Percentage of perfect matchings.

Figure 4: LTIU varying $p_2$ for different values of $p_1$.

We also consider the number of steps needed by our algorithm. From Figure 4(b), we can see that the number of steps is less than 2000 most of the time, except for problems with a large amount of incompleteness (i.e. $p_1$ = 0.8). As expected, with $p_1 > 0.6$ the algorithm requires more steps. In some cases, it reaches the step limit of 50000. Moreover, as the percentage of ties rises, stability becomes easier to achieve and thus the number of steps tends to decrease slightly. From the results we see that complete indifference ($p_2$=1) is a special case. In this situation, the number of steps increases for almost every value of $p_1$. This is because the algorithm makes most of its progress via random restarts. In these problems every person (if accepted) is equally preferred to all others accepted. The only blocking pairs are those involving singles who both accept each other. Hence, after a few steps all singles that can be married are matched, stability is reached, and the neighborhood becomes empty. The algorithm therefore randomly restarts. In this situation it is very difficult to find a perfect matching and the algorithm therefore often reached the step limit.

The algorithm is fast. It takes, on average, less than 40 seconds to give a result even for very difficult problems (see Figure 4(c)). As expected, with $p_2 = 1$ the time increases for the same reason discussed above concerning the number of steps.

Re-considering Figure 4(a) and the fact that all the marriages the algorithm finds are stable, we notice that most of the marriages are perfect. From Figure 4(d) we see that the average percentage of matchings that are perfect is almost always 100% and this percentage only decreases when the incompleteness is large. We compared our local search approach to the one in [4]. In their experiments, they measured the maximum size of the stable marriages in problems of size 10, fixing $p_1$ to 0.5 and varying $p_2$ in [0,1]. We did similar experiments, and obtained stable marriages of a very similar size to those reported in [4]. This means that although our algorithm is incomplete in principle, it always appears to find an optimal solution in practice, and for small sizes it behaves like a complete algorithm in terms of size of the returned marriage. However, it can also tackle problems of much larger sizes, still obtaining optimal solutions most of the times. We also considered the runtime behavior of our algorithm. In Figure 5(a) we show the average normalized number of blocking pairs and, in Figure 5(b), the average normalized number of singles of the best marriage as the execution proceeds. Although the step limit is 50000, we only plot results for the first steps because the rest is a long plateau that is not very interesting. We show the results only for $p_2 = 0.5$. However, for greater (resp., lower) number of ties the curves are shifted slightly down (resp., up). From Figure 5(a) we see that the average number of blocking pairs decreases very rapidly, reaching 5 blocking pairs after only 100 steps. Then, after 300-400 steps, we almost always reach a stable marriage, irrespective of the value of $p_1$. Considering Figure 5(b), we see that the algorithm starts with more singles for greater values of $p_1$. This happens because, with more incompleteness, it is more difficult for a person to be accepted. However, after 200 steps, the average number of singles becomes very small no matter the incompleteness in the problem.

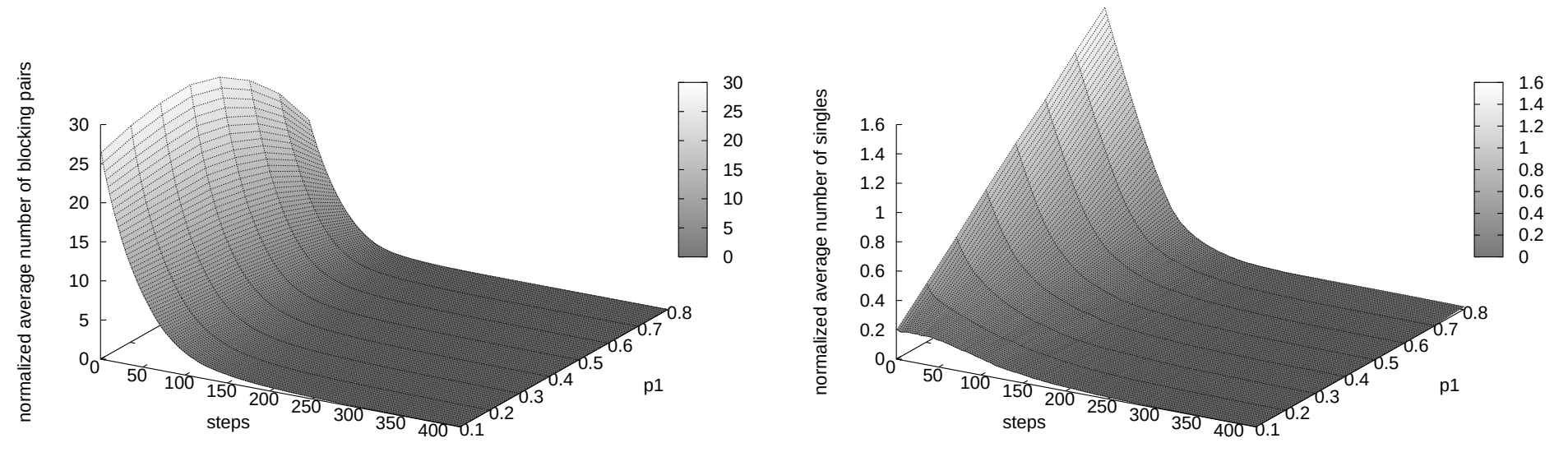

(a) Average normalized number of blocking pairs.

(b) Average normalized number of singles.

Figure 5: LTIU runtime behaviour ($p_2$=0.5).

Looking at both Figures 5(a) and 5(b), we observe that, although we set a step limit $s = 50000$, the algorithm reaches a very good solution after just 300-400 steps. After this number of steps, the best marriage found by the algorithm usually has no blocking pairs nor singles. This appears largely independent of the amount of incompleteness and the number of ties in the problems. Hence, for SMTI problems of size 100 we could set the step limit to just 400 steps and still be reasonably sure that the algorithm will return a stable marriage of a large size, no matter the amount of incompleteness and ties.

# 8 Conclusions and future works

We have presented a local search approach for solving the classical stable marriage (SM) problem and its variant with ties and incomplete lists (SMTI). Our algorithm for SM prob-lems has a simple scaling and size independent behavior and it is able to find a solution in a number of steps which grows as little as $O(n \log(n))$. Moreover it samples the stable marriage lattice reasonably well and it is a fair method to generate random stable marriages. We also provided an algorithm for SMTI problems which is both fast and effective at finding large stable marriages for problems of sizes not considered before in the literature. The algorithm was usually able to obtain a very good solution after a small amount of time.

We plan to apply a local search approach also to the hospital-resident problem and to compare our algorithms to the ones in [13], where residents express their preferences in strict order and hospitals allow ties in their preferences and have a finite number of posts each. We also aim to compare our algorithm with the Markov-chain-based model in [1] on the basis of execution time and sampling capabilities.

Mirco Gelain, Maria Silvia Pini, Francesca Rossi, Kristen Brent Venable
Department of Pure and Applied Mathematics, University of Padova, Italy
Email: {mgelain,mpini,frossi,kvenable}@math.unipd.it

Toby Walsh
NICTA and UNSW, Sidney, Australia
Email: toby.walsh@nicta.com.au